\documentclass[conference]{IEEEtran}
\IEEEoverridecommandlockouts

\usepackage{cite}
\usepackage{amsmath,amssymb,amsfonts,multirow}
\usepackage{booktabs}  
\usepackage{algorithmic}
\usepackage{graphicx}
\usepackage{textcomp}
\usepackage{xcolor}

\usepackage{epsfig,bbm,booktabs,fontawesome,bbding,arydshln,tabularx,multirow,colortbl,array,makecell}

\usepackage[hidelinks]{hyperref}

\RequirePackage{xspace}
\makeatletter
\DeclareRobustCommand\onedot{\futurelet\@let@token\@onedot}
\def\@onedot{\ifx\@let@token.\else.\null\fi\xspace}

\def\eg{\emph{e.g}\onedot} 
\def\ie{\emph{i.e}\onedot}

\def\etal{\emph{et al}\onedot}
\makeatother

\def\BibTeX{{\rm B\kern-.05em{\sc i\kern-.025em b}\kern-.08em
    T\kern-.1667em\lower.7ex\hbox{E}\kern-.125emX}}
\begin{document}

\title{Image Demoiréing Using Dual Camera Fusion on Mobile Phones}

\author{
	Yanting Mei, Zhilu Zhang, Xiaohe Wu$^{\dagger}$, Wangmeng Zuo \\ 
    Faculty of Computing, Harbin Institute of Technology, Harbin, China \\
    meiyt0418@gmail.com, cszlzhang@outlook.com, 
    csxhwu@gmail.com, wmzuo@hit.edu.cn
}
\maketitle
\renewcommand{\thefootnote}{}

\footnotetext{$^{\dagger}$ Corresponding Author. }
\footnotetext{This work was supported in part by the National Natural Science Foundation of China (NSFC) under Grant No.62476067, Heilongjiang Science and Technology Project under Grant No.2022ZX01A21.}

\begin{abstract}
When shooting electronic screens, moiré patterns usually appear in captured images, which seriously affects the image quality. Existing image demoiréing methods face great challenges in removing large and heavy moiré. To address the issue, we propose to utilize \textbf{D}ual \textbf{C}amera fusion for  \textbf{I}mage \textbf{D}emoiréing (DCID), \ie, using the ultra-wide-angle (UW) image to assist the moiré removal of wide-angle (W) image. This is inspired by two motivations: (1) the two lenses are commonly equipped with modern smartphones, (2) the UW image generally can provide normal colors and textures when moiré exists in the W image mainly due to their different focal lengths. In particular, we propose an efficient DCID method, where a lightweight UW image encoder is integrated into an existing demoiréing network and a fast two-stage image alignment manner is present. Moreover, we construct a large-scale real-world dataset with diverse mobile phones and monitors, containing about 9,000 samples. Experiments on the dataset show our method performs better than state-of-the-art methods. Code and dataset are available at \url{https://github.com/Mrduckk/DCID}.

\end{abstract}


\section{Introduction}
\label{sec:intro}

It has become a common and convenient way to use smartphones to record and transfer data from electronic screens. However, Moiré appears in the captured image, presented as wavy, concentric, and other complex geometric patterns, while it does not exist in the original image.
This occurs due to the frequency aliasing phenomenon between the pixel array of the camera sensor and the pixel grid of the displayed screen. Moiré not only affects visual quality, but also may lead to information loss of image textures and details. Therefore, the technique of removing moiré has received widespread attention.

\begin{figure}[t]
    \centerline{\includegraphics[width=1\linewidth]{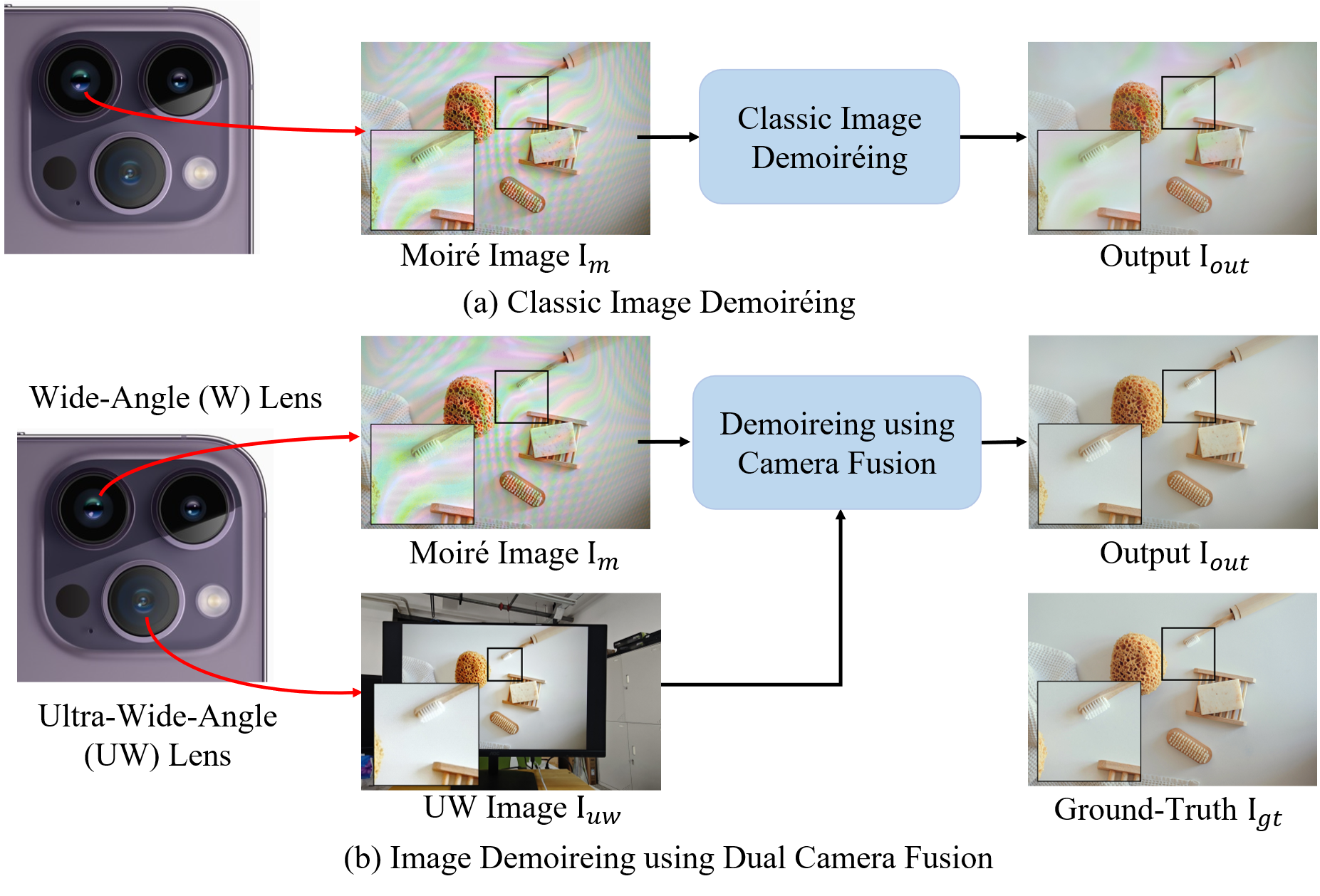}}
    \vspace{-4mm}
    \caption{Comparison of image demoiréing methods. The sub-figure (a) shows a classic image demoiréing with a single input. The sub-figure (b) shows image demoiréing using asymmetric camera fusion.}
    \label{intro}
\vspace{-6mm}
\end{figure}

Early moiré removal methods adopt traditional machine learning techniques such as low-rank sparse matrix decomposition~\cite{liu2015moire} and bandpass filtering~\cite{yang2017textured}. With the rise of deep learning, many works design neural networks. For instance, MopNet~\cite{he2019mop} introduces moiré pattern classification to process them in a divide-and-conquer manner. MBCNN~\cite{zheng2020image} integrates frequency domain modeling. FHDe$^2$Net~\cite{he2020fhde} and ESDNET~\cite{yu2022towards} improve multi-scale processing architecture, experimenting successfully on high-definition and ultra-high-definition (UHD) images, respectively. Although these methods have made great progress in removing light and small moiré patterns, they do not have satisfactory performance on severe moiré ones, as shown in  Fig.~\ref{intro} (a). One possible solution is to design a complex and large network to increase model fitting ability. But it would inevitably bring greater computational costs, leading to time-consuming or infeasible inference of UHD images on mobile phones.

To address the issue efficiently, we propose \textbf{D}ual \textbf{C}amera fusion for  \textbf{I}mage \textbf{D}emoiréing (DCID), \ie, using the ultra-wide-angle (UW) image to assist the moiré removal of wide-angle (W) image. It is based on two motivations. First, modern smartphones are generally equipped with asymmetric multi-lens camera systems, making it possible to collect images from both wide-angle and ultra-wide-angle cameras. Second, from our observation, moiré is sensitive to lens focal length, distance, and angle between the camera and display. W lens serves as the commonly used camera for capturing screen-display contents. When severe frequency aliasing exists between the W sensor and the screen subpixel, the UW image usually does not present obvious moiré patterns, due to the different focal lengths and positions of the UW and W lenses. Thus, the UW image can provide information with low resolution but relatively normal textures and colors, improving the performance for W image demoiréing, as shown in Fig.~\ref{intro} (b).

In particular, we propose an efficient DCID method.
We take a recent single image demoiréing network (\ie, ESDNet~\cite{yu2022towards}) as backbone, and design a lightweight encoder for the UW image. Spatial alignment between UW and W is crucial. Some methods like optical flow~\cite{huang2022flowformer} and deformable convolution~\cite{dai2017deformable} prove ineffective or computationally costly. To resolve this, we propose a fast and efficient two-stage alignment method, first applying alignment on the image level by keypoint matching and then doing that on the feature level by convolution kernel prediction. Finally, the aligned features from UW and W encoders are adaptively fused and fed into the decoder.

Moreover, we construct a large-scale real-world DCID dataset.
During the data collection, we focus more on moiré images with messy colors and complex patterns, as this moiré is more difficult to remove and the previous methods generally perform poor ability on it. Using 3 mobile phones to capture 3 displays successively, we obtaining about 9,000 samples. During the data pre-processing, we crop the captured UW image and moiré W image as input, and align the source image to the cropped W image as the ground-truth. Moreover, display settings (\eg, brightness, contrast, and color adjustment parameters) make different color tones between the source and captured images, thus, we further align the brightness and color of the source image to the cropped W image ones.

Experiments are conducted on our DCID dataset. The results show that our DCID method achieves better than state-of-the-art image demoiréing methods both in quantitative and qualitative comparison, which demonstrates the effectiveness of the dual camera fusion manner for moiré removal.

Our main contributions can be summarized as follows:

\begin{itemize}
	\item To improve the effect of severe moiré removal on mobile phones, we propose to utilize \textbf{D}ual \textbf{C}amera fusion for \textbf{I}mage \textbf{D}emoiréing (DCID), where the ultra-wide-angle (UW) image is taken to assist wide-angle (W) image demoiréing.  
	\item We integrate a lightweight UW image encoder into an existing demoiréing network, and propose a fast and efficient two-stage manner to align UW and W images.
	\item We construct a large-scale real-world DCID dataset with diverse smartphones and monitors, consisting of about 9,000 samples in total. 
       \item Experiments on DCID dataset show our method behaves favorably against state-of-the-art image demoiréing methods both quantitatively and qualitatively.
\end{itemize}

\section{Related Work}

\subsection{Image demoiréing}

With the rise of mobile photography, moiré patterns have become prevalent, posing significant challenges for image restoration. Traditional demoiréing methods, such as low-rank decomposition~\cite{liu2015moire} and layer decomposition on polyphase components (LDPC)~\cite{yang2017textured}, focus on signal processing but incur high computational costs. Deep learning has revolutionized image demoiréing. Early CNN-based approaches, such as MopNet~\cite{he2019mop} and MBCNN~\cite{zheng2020image}, leveraged frequency domain optimization and texture decomposition. Sun~\etal~\cite{sun2018moire} proposed a multi-scale learning convolutional neural network and the first benchmark dataset captured on LCD screens. Two-stage architectures like FHDe$^2$Net~\cite{he2020fhde} introduced global-to-local strategies for better texture and color preservation. More recent methods, such as ESDNet~\cite{yu2022towards}, focus on lightweight designs for ultra-high-definition image moiré removal. MCFNet~\cite{nguyen2023multiscale} proposed a multiscale guided screenshot demoiréing method to understand moiré frequency correlation.

Although significant progress has been made, these methods are limited by the characteristics of the single image itself and cannot remove severe moiré patterns. Considering the current UHD images captured by mobile cameras, we propose to utilize Dual Camera fusion for Image Demoiréing (DCID), where the UW image is taken to assist the W image demoiréing.

\subsection{Reference-based Image Restoration}

Reference-based image processing idea is applied in multiple image restoration tasks. For example, RefSR~\cite{zhang2022self1, zhang2024self,shim2020robust} leverages high-resolution reference images that share similar content and textures to assist the super-resolution of low-resolution input.
Zhang~\etal~\cite{zhang2022self2, zhang2024bracketing} utilizes blurry long-exposure images to help denoising of sharp short-exposure images.
Liu~\etal~\cite{liu2020self} introduced a self-adaptive learning method, utilizing an additional defocused moiré-free image to assist in removing moiré patterns from the focused moiré image. 

Modern smartphones make UW images easily accessible, and UW images offer richer and detailed information than defocused images. To the best of our knowledge, this work is the first dual-camera fusion method for image demoiréing.

\section{Dataset}
\begin{table*}[t]
\caption{Mobile phone cameras and digital screens used in our dataset.}

\vspace{-4mm}
\begin{center}
\begin{tabular}{ccccccc}
\toprule
\multicolumn{4}{c}{Mobile Phone}& \multicolumn{3}{c}{Digital Screen} \\
\cmidrule(lr){1-4}\cmidrule(lr){5-7}
Manufacturer&Model&W Image Size& UW Image Size& Manufacturer&Model&Resolution\\
\midrule
HUAWEI& P40 & $4096\times3072$&$4608\times3456$ & REDMI & RMMNT238NF &$1920\times1080$\\
Xiaomi&14&$4096\times3072$ &$4080\times3060$& AOC &24B1XH5  & $1920\times1080$\\
iPhone&14 Pro&$4032\times3024$ &$4032\times3024$ &KTC & H27T13 & $2560\times1440$\\
\bottomrule
\end{tabular}
\label{dataset}
\end{center}
\vspace{-6mm}
\end{table*}
To achieve DCID, we collect a large-scale real-world dataset using various smartphones and monitors. Below, we detail the dataset construction process.

\subsection{Data Collection}

Each sample in the dataset should contain three images, \ie, moiré W images, UW images, and ground-truth (GT). The moiré W and UW images (with less moiré) can be captured using smartphones, where UW images serve as valuable auxiliary information for W image demoiréing. The moiré-free GT image can be obtained from the associated source image displayed on the screen.

Our dataset contains a wide range of common screen display scenarios, including natural scene images, web content, and other materials such as documents and research papers. Moiré patterns with messy colors and intricate geometry significantly affect visual perception and are hard to remove using previous methods. Therefore, we are more concerned with collecting images under these moiré patterns.  
Moreover, to collect diverse moiré patterns, we capture images with various angles, distances, and lighting conditions, which produce moiré images with a wide array of shapes, scales, and color characteristics.

We use three different brands of smartphones to shoot three different types of displays, which form 9 (3$\times$3) combinations. The detailed configuration is provided in Table~\ref{dataset}. About 1,000 samples are taken for each combination. Finally, there are 8959 samples in total.
The training and test sets consist of 7,176 and 1,783 samples, respectively. 
Each combination contains $\sim$200 test samples.
Compared to existing datasets~\cite{sun2018moire,yuan2019aim,he2020fhde,yu2022towards}, the images in this dataset exhibit more severe moiré patterns. It significantly increases the difficulty of restoring moiré-free content, providing a challenging benchmark for future works.

\begin{figure}[t]
\centerline{\includegraphics[width=1\linewidth]{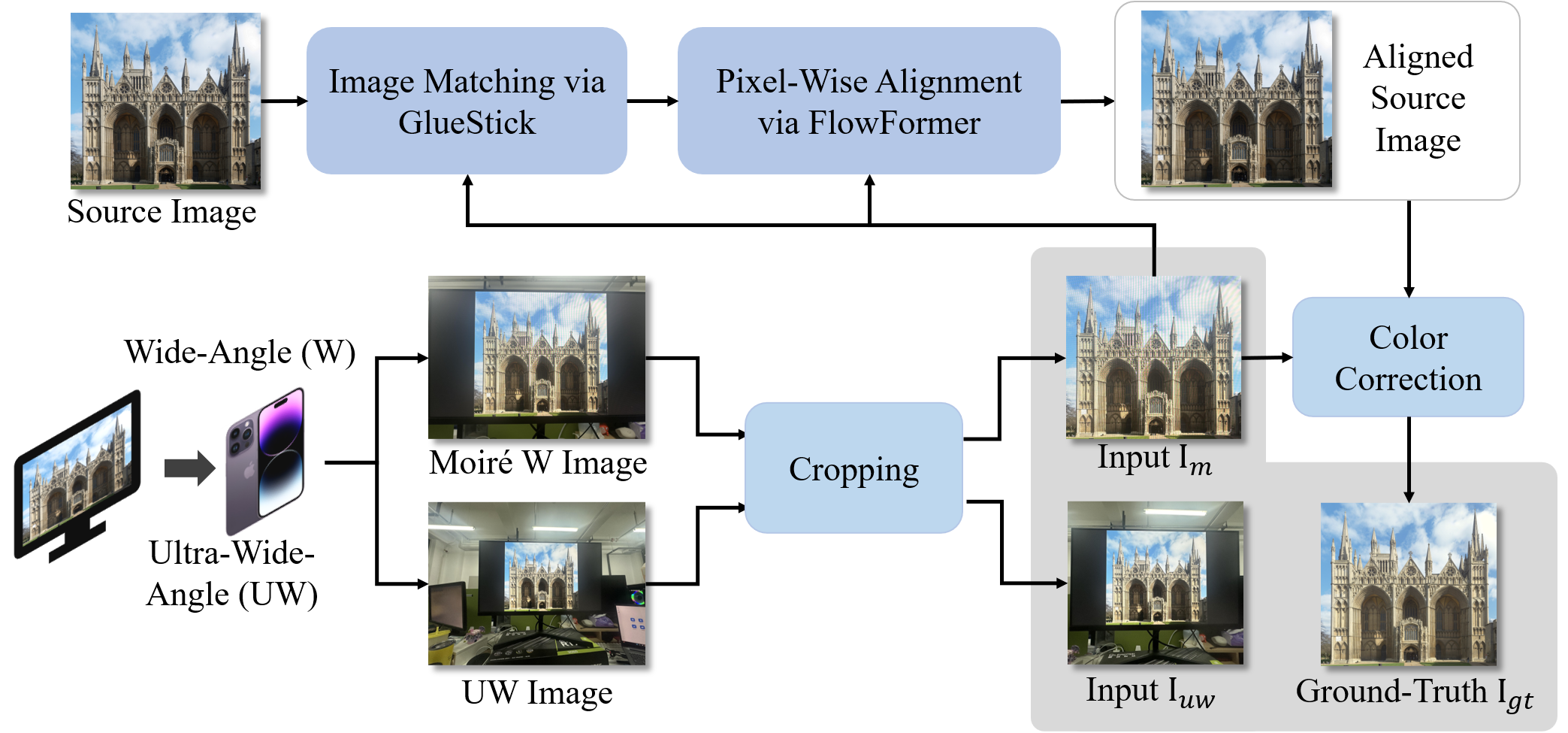}}
\vspace{-3mm}
\caption{Data processing pipeline in our DCID dataset.}
\label{dataset_pipeline}
\vspace{-5mm}
\end{figure}

\subsection{Data Processing}

We first crop the main screen content from the W image, getting the moiré input $I_m$. Then the UW image is cropped with the same cropping ratio, getting the auxiliary input $I_{uw}$.
To better align $I_m$ and source image $I_{gt}^{src}$ spatially, we perform a two-step image alignment. First, we use Gluestick~\cite{pautrat2023gluestick} to perform image matching based on keypoints and keylines, thus roughly aligning the source image to the cropped W image. Next, we further apply a pixel-wise alignment using FlowFormer~\cite{huang2022flowformer}, to align the initially-aligned source image to $I_m$, getting the finally-aligned source image $\Tilde{I}_{gt}^{src}$.

In existing image demoiréing datasets~\cite{sun2018moire,yuan2019aim,he2020fhde,yu2022towards}, the brightness and color of GT are generally the same as those of the source image. However, it may not be appropriate, as different display settings (\eg, brightness, contrast, and color adjustment parameters) bring various displayed image appearances. 
To ensure that the demoiréing method only removes the moiré and does not alter the color representation of the captured image, we perform additional color correction on $\Tilde{I}_{gt}^{src}$. Specifically, a $3\times3$ color correction matrix is estimated using the least squares method, which represents the linear color mapping between $I_m$ and $\Tilde{I}_{gt}^{src}$. Note that the matrix is calculated between Gaussian blurred $I_m$ and $\Tilde{I}_{gt}^{src}$, which reduces the impact of moiré in $I_m$ and helps to achieve a more accurate color mapping. Finally, we apply the matrix to $\Tilde{I}_{gt}^{src}$, getting the GT $I_{gt}$. $I_{gt}$ can more faithfully reflect the color state of the scene displayed on the screen.

\section{Method}

In this section, we introduce our proposed DCID method. We first revisit ESDNet~\cite{zhang2018residual}. Then we describe the overall pipeline of DCID method and detail the two-stage alignment manner. Finally, we give the loss functions.

\subsection{Revisiting ESDNet}

ESDNet~\cite{zhang2018residual} is a multi-scale encoder-decoder architecture, with a specially designed block at each scale.
Each ESDNet~\cite{zhang2018residual} block contains a dilated residual dense module and a plug-and-play semantic alignment scale-aware module (SAM). SAM consists of two main modules, \ie, a pyramid feature extraction module and a cross-scale dynamic fusion module, which effectively capture multi-scale features and dynamically integrate image features by learning fusion weights, respectively. These innovations enable ESDNet to achieve superior performance compared to previous works on the LCDmoiré~\cite{yuan2019aim}, FHDMi~\cite{he2020fhde}, and UHDM~\cite{yu2022towards} datasets.

\subsection{Method Overview}

When facing complex and irregular moiré patterns, some single image demoiréing methods, \eg, ESDNet, struggle to remove them. To solve this problem, we propose Dual Camera Fusion for Image Demoiréing (DCID), \ie, using the UW image to assist the moiré removal of the W image. UW images, easily captured with smartphones, offer additional information with lower resolution but relatively normal textures, thereby enhancing the performance of W image demoiréing.

\begin{figure}[t]
\centerline{\includegraphics[width=1\linewidth]{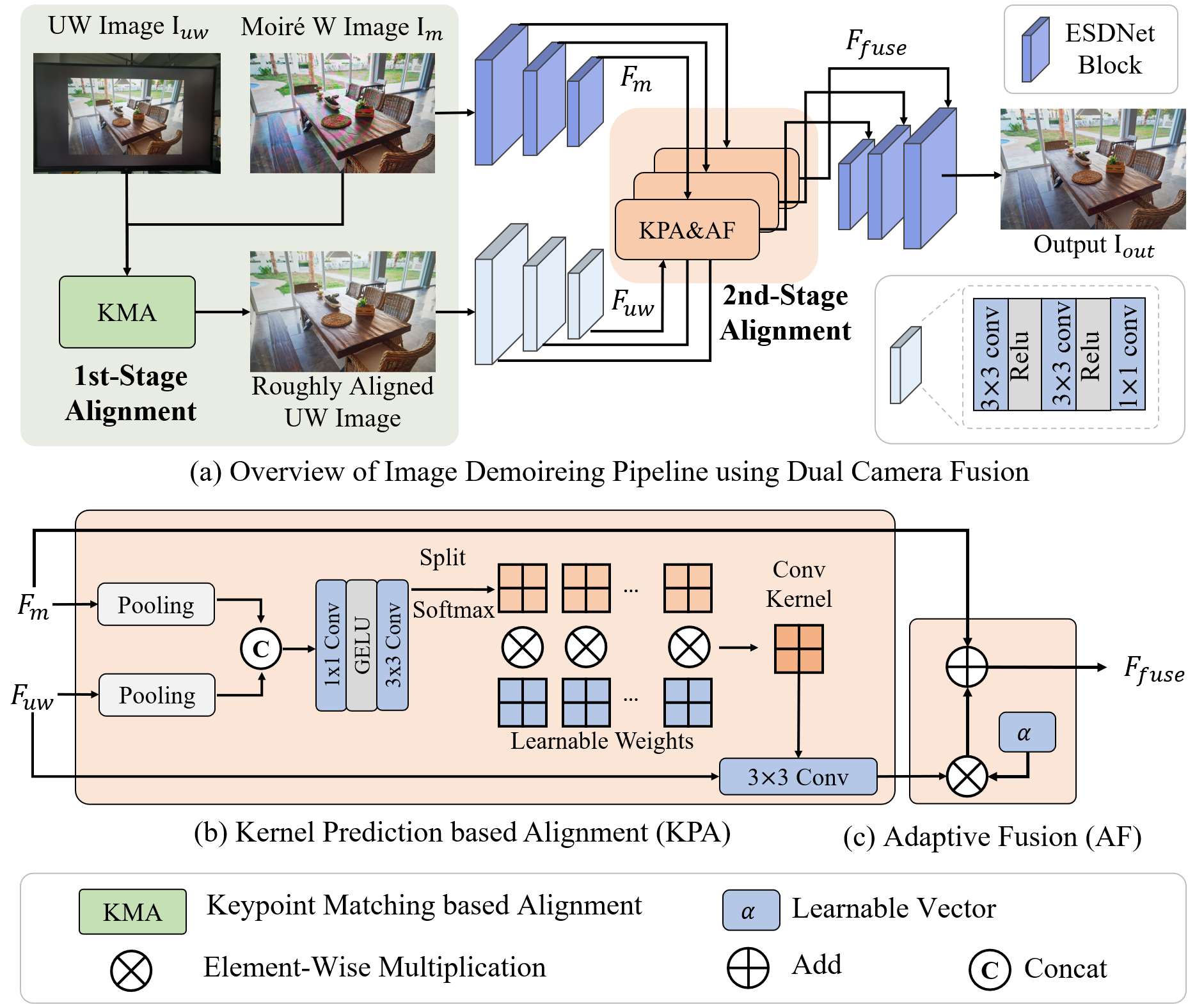}}
\vspace{-3mm}
\caption{Pipeline of the proposed DCID method.}
\label{model_pipeline}
\vspace{-5mm}
\end{figure}

The overall architecture of our DCID method is illustrated in Fig.~\ref{model_pipeline}. We take ESDNet as the backbone and design a lightweight encoder for the UW image to integrate into it. The lightweight encoder extracts the feature from the UW image, and adopts a simple architecture consisting of three convolutional blocks, each of which contains a downsampling layer and 3-layer convolutions.
Compared with ESDNet, this design brings a small amount of additional computational costs.

Then what needs to be done is the alignment and fusion between the UW and W images. Spatial alignment is crucial for getting satisfactory results, but existing alignment manners are generally computationally intensive and time-consuming. In this work, we propose a fast and efficient two-stage alignment method to balance efficiency and performance, which is described in detail in the next subsection. Finally, the aligned features from W and UW encoders are adaptively fused and fed into the decoder.

\subsection{Two-Stage Alignment} 
The proposed alignment framework consists of two stages, \ie, Keypoint Matching based Alignment (KMA) and Kernel Prediction based Alignment (KPA). The first stage performs coarse geometric alignment on the image level, and the second performs fine-grained alignment on the feature level.

\textbf{KMA}. There is a significant difference in the field of view between UW and W images. In this case, the common optical flow based alignment methods are not cost-effective, \ie, achieving a good alignment performance may come at a high cost.
Instead, we propose to perform sparse keypoint matching between the two images, and then utilize the matching prompts to align them. Specifically, we apply SuperPoint~\cite{detone2018superpoint} to detect 256 keypoints from the $\times4$ down-sampled images, and then adopt   LightGlue~\cite{lindenberger2023lightglue} to match the keypoints. The keypoints number and down-sampling factor are determined through extensive experiments, and these settings can provide a better trade-off between computational efficiency and alignment performance.
Finally, the UW image $I_{uw}$ is roughly aligned to W image $I_{m}$ according to the keypoint matching results, \ie,
\begin{equation}
\hat{I}_{uw} = \text{KMA}(I_m,I_{uw}),
\end{equation}
where $\hat{I}_{uw}$ denotes the aligned UW image.

\begin{figure*}[htbp]
\centerline{\includegraphics[width=1.0\textwidth]{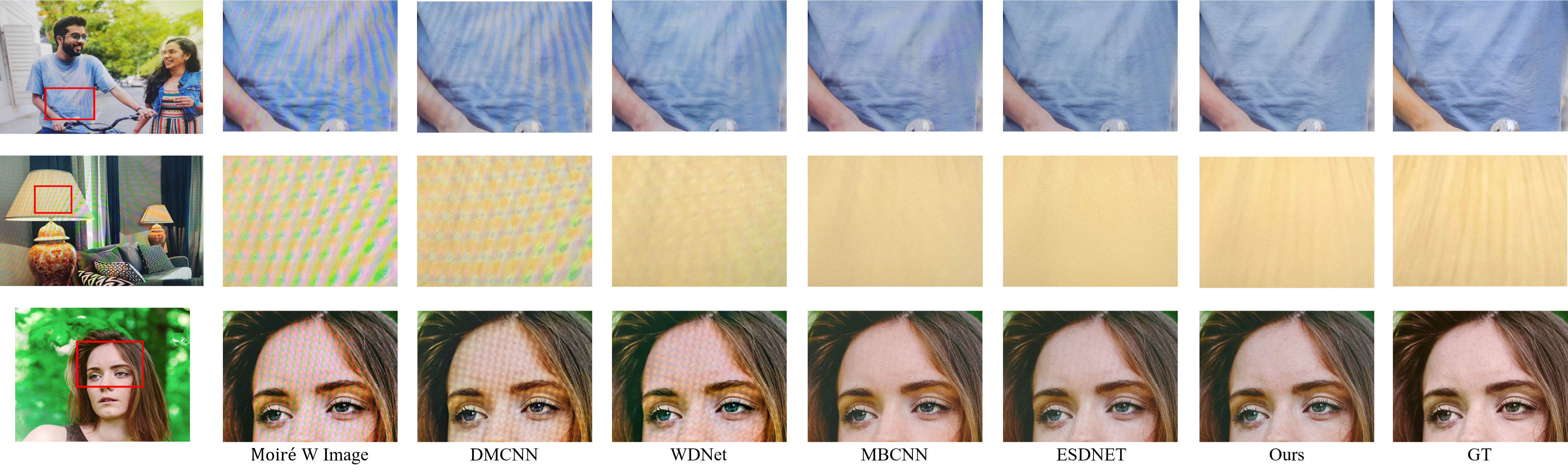}}
\vspace{-5mm}
\caption{Visual comparison of demoiréing results among state-of-the-art methods on our dataset. Please zoom in for a better view.}
\label{output}
\vspace{-6mm}
\end{figure*}

\textbf{KPA}. In the second stage, the KPA module further aligns $\hat{I}_{uw}$  and $I_{m}$ spatially. We propose to perform that on the multi-scale feature level, as multi-scale features may integrate more fine-grained information and alleviate the adverse impact of moiré on alignment. The W features $F_m\in\mathbb{R}^{C \times H \times W}$ and UW features $F_{uw} \in \mathbb{R}^{C \times H \times W}$ are extracted through their encoders. 
Applying deformable convolution~\cite{dai2017deformable} is a possible choice, but we found that it does not work well here, probably due to unstable optimization.
Instead, we estimate transform convolution weights for $F_{uw}$. TransXNet~\cite{lou2023transxnet}, we construct a KPA module to achieve this.
Specifically, adaptive average pooling is first adopted to aggregate contextual information of $F_m$ and $F_{uw}$, and then they are concatenated along the channel dimension, getting $W'$, \ie,
\begin{equation}
W' = \text{Concat}(\text{AdaptivePool}(F_m), \text{AdaptivePool}(F_{uw})).
\end{equation}
Next, two convolutional layers are applied to process $W'$. To generate the initial convolution kernel weights $W \in \mathbb{R}^{G \times C \times K^2}$ (where $G$ is the number of weight groups and $K$ is the convolution kernel size), we use a softmax function along the group dimensions, \ie,
\begin{equation}
W = \text{Softmax}(\text{Reshape}(\text{Conv}(\text{ReLU}(\text{Conv}(W'))))).
\end{equation}
By performing element-wise multiplication between $W$ and learnable parameters $P$, followed by summation along the group dimension, we can obtain a content-adaptive convolution kernel $\theta \in \mathbb{R}^{C \times K \times K}$, \ie,
\begin{equation}
\theta = \sum_{i=0}^{G} P_i W_i .
\end{equation}
This kernel is used to transform $F_{uw}$ into $\tilde{F}_{uw}$, \ie,
\begin{equation}
\tilde{F}_{uw} = \text{Conv}(F_{uw}; \theta).
\end{equation}
$\tilde{F}_{uw}$ can be regarded as more fine alignment results from $\hat{I}_{uw}$ to $I_{m}$.

Finally, the fused feature $F_{fuse}$ is then computed by adaptively combining $F_m$ and $\tilde{F}_{uw}$ using a learnable coefficient $\alpha \in \mathbb{R}^{C \times 1 \times 1}$, $\alpha$ is adaptively learned during training,  \ie, 
\begin{equation}
F_{fuse} = F_m + \alpha  \tilde{F}_{uw}.
\end{equation}
This fusion strategy effectively integrates information from both W and UW features, ensuring that essential structural and contextual details in W features are preserved.

\subsection{Loss Functions}
 To enhance the network's capability to remove moiré patterns across multiple scales, we employ a multi-scale loss function. It is applied by a multi-scale supervision mechanism, which constrains the network to learn consistent features across different spatial scales. Moreover, the total loss function integrates $\ell_1$ and perceptual loss $\mathcal{L}_{p}$ terms, which  can be written as,
\begin{equation}
\mathcal{L}_{\text{total}}=\sum_{i=1}^3( \lVert I_{out}^i - I_{gt}^i \rVert_1 + \lambda_p \lVert \phi(I_{out}^i) - \phi(I_{gt}^i) \rVert_1 ).
\end{equation}
where $\phi(\cdot)$ represents the features extracted by the VGG~\cite{simonyan2014very}  network, $\lambda_p$ is the weight of perceptual loss, which is set to 2.0, $I_{out}^i$ represents the predicted outputs at scale $i$, and $I_{gt}^i$ is $\times 2^{i-1}$ downsampled version of the original ground truth $I_{gt}$.

\begin{table}[t]
\centering
\footnotesize
\caption{Quantitative comparisons between our model and state-of-the-art methods on our datasets. $\uparrow$ denotes the larger the better, and $\downarrow$ denotes the smaller the better. Methods marked in \textbf{bold} and \underline{underlined} indicate the best and the second best one, respectively.} 
\vspace{-2mm}
\label{output_table}
\begin{tabular}{c|c|cccc}
\toprule
Dataset & Methods & PSNR$\uparrow$ & SSIM $\uparrow$ & LPIPS $\downarrow$ & $\Delta$E $\downarrow$ \\
\midrule
\multirow{6}{*}{Xiaomi} 
& DMCNN~\cite{sun2018moire} & 23.36& 0.8586&0.314 &6.308 \\
& WDNet\cite{liu2020wavelet} &25.66 &0.8699 &0.239 &4.670 \\
& MBCNN\cite{zheng2020image} &25.44 &0.8826 &0.213 &4.496 \\
& ESDNet\cite{yu2022towards} & \underline{26.14}& \underline{0.8889}& \underline{0.204}& \underline{4.166}\\
& Ours &\textbf{27.06} &\textbf{0.8973} &\textbf{0.197} &\textbf{3.777} \\
\midrule
\multirow{6}{*}{HUAWEI} 
& DMCNN\cite{sun2018moire} &25.12  &0.8653 &0.338 &5.449 \\
& WDNet\cite{liu2020wavelet} &26.69 &0.8734 & 0.252&4.383 \\
& MBCNN\cite{zheng2020image} &26.23 &0.8805 &0.226 &4.339 \\
& ESDNet\cite{yu2022towards} & \underline{27.31}& \underline{0.8864}&\underline{0.220} &\underline{3.892} \\
& Ours &\textbf{27.72} &\textbf{0.8926} & \textbf{0.212} &\textbf{3.783} \\
\midrule
\multirow{6}{*}{iPhone} 
& DMCNN\cite{sun2018moire} &24.58 & 0.8617&0.333  & 5.473\\
& WDNet\cite{liu2020wavelet} & 26.08&0.8735 &0.259 & 4.316\\
& MBCNN\cite{zheng2020image} &25.97 &0.8806 & 0.242 & 4.158\\
& ESDNet\cite{yu2022towards} & \underline{26.36}& \underline{0.8827}& \underline{0.232}& \underline{3.933}\\
& Ours &\textbf{26.99} & \textbf{0.8831}&\textbf{0.228} & \textbf{3.686}\\
\bottomrule
\end{tabular}

\vspace{-4mm}
\end{table}

\section{experiment}
\subsection{Experimental Settings}
We evaluate our method on the proposed dataset. To quantitatively assess performance, we employ widely used metrics, including PSNR, SSIM~\cite{wang2004image}, LPIPS~\cite{zhang2018unreasonable}, and $\Delta$E. PSNR and SSIM are used to measure pixel-wise fidelity and structural similarity, while LPIPS provides a perceptual quality metric aligned with human visual perception. These metrics are widely used in image restoration tasks, including the previous demoiréing works such as ESDNet~\cite{yu2022towards} and MBCNN~\cite{zheng2020image}. 
$\Delta$E is used to quantify the perceptual color difference between restored and ground-truth images, as chaotic colors often exist in moiré patterns. This metric is also adopted in the recent image demoiréing method~\cite{zhang2023real}. 

Our model was implemented using PyTorch on an NVIDIA RTX A6000 GPU. During training, a batch size of 8 is used with 768$\times$768 randomly cropped patches. Initial learning rate 0.0002 follows cyclic cosine annealing~\cite{loshchilov2016sgdr}. Optimization is performed using the Adam optimizer~\cite{kingma2014adam}, with $\beta_1$ = 0.9 and $\beta_2$ = 0.999. The model is trained for 500 epochs. For evaluation, the model is tested on full-resolution images without resizing, reflecting real-world application scenarios. To ensure fair evaluation, we also trained competing methods on our dataset using the same configurations.

\subsection{Comparison with the Previous Works}

\textbf{Quantitative Comparison}. 
We compare our method against state-of-the-art approaches, including DMCNN~\cite{sun2018moire}, WDNet~\cite{liu2020wavelet} MBCNN~\cite{zheng2020image}, and ESDNet~\cite{yu2022towards}. These methods only input the moiré image. Table~\ref{output_table} presents the quantitative results on our dataset. Our method achieves state-of-the-art performance, significantly outperforming existing methods across all metrics. Taking the results on the Xiaomi camera dataset as an example, our method improves PSNR by 0.92dB and $\Delta$E by 0.389 in comparison with ESDNet. It shows the UW image can provide valuable information and help to remove moiré patterns effectively. 
Additionally, some recent methods do not directly applicable to our dataset. For example, RRID~\cite{xu2024image} operates on both RAW and RGB images, and UnDeM~\cite{zhong2024learning} uses unpaid data for training, while our method focuses on supervised demoiréing on RGB images. To demonstrate the effectiveness of our methods, we adapt them to our settings to conduct experiments. For RRID, we replace the RAW input with RGB image. For UnDeM, we use it to generate pseudo-moiré images and take them to train the demoiréing model. From Table~\ref{table_rrid}, our method still outperforms the two methods. 
Nevertheless, since our method builds upon ESDNet as a baseline, we place particular emphasis on comparing our results with ESDNet. 

\begin{table}[t]
\centering
\footnotesize
\caption{Other method comparisons on Xiaomi camera dataset.} 
\vspace{-2mm}
\label{table_rrid}
\begin{tabular}{c|c|cccc}
\toprule
Methods & UW & PSNR$\uparrow$ & SSIM$\uparrow$ &
        LPIPS$\downarrow$ & $\Delta$E$\downarrow$  \\
       \midrule
        RRID~\cite{xu2024image} & × & 25.47 & 0.8705 & 0.233 & 4.474 \\
        UnDeM~\cite{zhong2024learning} & × & 23.73 & 0.8531 & 0.246 & 6.046 \\
        \midrule
        Ours + RRID~\cite{xu2024image} & \checkmark & 26.01 & 0.8781 & 0.214 & 4.282 \\
        Ours + UnDeM~\cite{zhong2024learning} & \checkmark  & 24.51 & 0.8670 & 0.222&5.314 \\
\bottomrule
\end{tabular}
\vspace{-2em}
\end{table}

\textbf{Qualitative Comparison}. Fig.~\ref{output} illustrates the visual comparison between our method and existing methods. Our method consistently delivers more perceptually satisfying results. By incorporating UW images, our approach effectively addresses challenges posed by both large-scale and small-scale textures, significantly alleviating the issues caused by moiré patterns. For instance, in the second example in Fig.~\ref{output}, the textures on the lampshade are obscured by moiré interference in the moiré image, making it hard to restore the details using existing methods. In contrast, with the inclusion of UW images, our method successfully recovers these textures, demonstrating its ability to utilize auxiliary information to enhance the reconstruction of fine contents. 
This demonstrates the robustness and effectiveness of our proposed method, particularly in processing challenging scenarios.

\begin{table}[t]
\centering
\footnotesize
\caption{Ablation study of two-stage image alignment manners on Xiaomi camera dataset.} 
\vspace{-2mm}
\label{12align}
\setlength{\tabcolsep}{2pt}
\begin{tabular}{cc|ccc}
\toprule
1-Stage & 2-Stage & PSNR$\uparrow$ / SSIM$\uparrow$ / 
 LPIPS$\downarrow$ / $\Delta$E$\downarrow$ & 
\#FLOPs (G) &Time (s) \\
\hline
-&-&26.14 / 0.8889 / 0.204 / 4.166&4.478&0.518\\
OF  & -  &26.84 / 0.8933 / 0.200 / 3.928&7.698&1.219\\
KMA & - &26.93 / 0.8936 / \underline{0.198} / 3.882&5.020&0.618\\
OF  & DCN  &26.84 / 0.8931 / 0.200 / 3.910&9.940&1.414\\
KMA & DCN &26.85 / 0.8934 / 0.201 / 3.965&7.262&0.827\\
OF  & KPA  & \underline{26.94} / \underline{0.8944} / 0.199 / \underline{3.859}&7.700&1.250\\
KMA & KPA &\textbf{27.06} / \textbf{0.8973} / \textbf{0.197} / \textbf{3.777}&5.022&0.659\\

\bottomrule
\end{tabular}
\vspace{-5mm}
\end{table}

\subsection{Ablation Study}

To validate the effectiveness of our two-stage alignment manner, we conducted ablation studies on the Xiaomi camera dataset. The first line in Table~\ref{12align} shows the result that the UW image is not used, and we take it as the baseline. First, the single-stage alignment methods, whether based on optical flow (OF)~\cite{huang2022flowformer} or our KMA, exhibit limited metric improvement, indicating their unsatisfactory image alignment capabilities. In particular, OF doubles the inference time, which is obviously not cost-effective. Second, for the second stage alignment, we compared deformable convolution (DCN)~\cite{dai2017deformable} and the proposed KPA module. However, DCN does not bring performance gains, probably due to the difficulty of its optimization. In both OF-based and KMA-based methods, replacing DCN with KPA consistently achieved better PSNR and lower $\Delta$E, highlighting KPA’s efficiency and effectiveness in aligning features and suppressing moiré patterns. 
Overall, our two-stage alignment (KMA + KPA) design significantly improves the results while only introducing a small increase in computational cost.

We have conducted experiments on the perceptual loss coefficient $\lambda_p$. The results in Table~\ref{ablation_p} demonstrate that the perceptual loss significantly improves performance. When $\lambda_p$ is set to $ 2.0$, the model performs best.

\begin{table}[t]
\centering
\footnotesize
\caption{Ablation study of $\lambda_p$ on Xiaomi camera dataset.} 
\vspace{-2mm}
\label{ablation_p}
\setlength{\tabcolsep}{1.5pt}
\begin{tabular}{c|ccccc}
\toprule
$\lambda_p$ & 0&1&2&3&4\\
\hline
PSNR/SSIM&26.06/0.886&\underline{27.02}/\underline{0.895}&\textbf{27.06}/\textbf{0.897}&26.55 /0.893&26.45/0.893\\
 LPIPS/$\Delta$E&0.244/4.184&\underline{0.199}/\underline{3.843} &\textbf{0.197}/\textbf{3.777}&0.202/4.035&0.204/4.099 \\

\bottomrule
\end{tabular}
\vspace{-6mm}
\end{table}

\section{Conclusion}
In this paper, to remove the severe moiré, we propose to perform Dual-Camera fusion for Image Demoiréing (DCID), \ie, leveraging ultra-wide-angle (UW) images to assist wide-angle (W) image demoiréing. To support this, we construct a large-scale real-world DCID dataset with a diverse range of smartphones and monitors. Moreover, the proposed efficient DCID framework integrates a lightweight UW image encoder into the existing demoiréing network and introduces a fast two-stage image alignment manner. Experimental results on the DCID dataset demonstrate that our method significantly outperforms state-of-the-art ones in both quantitative and qualitative evaluations.

\bibliographystyle{IEEEbib}
\bibliography{main}

\end{document}